\documentclass{article}

\usepackage{PRIMEarxiv}
\usepackage{natbib}
\usepackage{color,soul}

\usepackage{import}
\usepackage{gensymb}

\usepackage{caption}
\usepackage{csvsimple}
\usepackage{booktabs, wrapfig}
\usepackage{array, ltablex, multirow}
\usepackage{placeins}
\usepackage{graphicx}
\usepackage{xcolor}
\usepackage{dblfloatfix} 
\usepackage{transparent}
\usepackage{subcaption}
\usepackage{mathrsfs}
\usepackage{derivative}
\usepackage{pgfplots}
\usepackage{csvsimple}
\usepackage{amsmath,amsfonts,amssymb,amsthm, bm}
\usepackage{svg}
\usepackage{pgfplots}
\usepgfplotslibrary{fillbetween}
\usepackage[utf8]{inputenc} 
\usepackage[T1]{fontenc}    
\usepackage{hyperref}       
\usepackage{url}            
\usepackage{booktabs}       
\usepackage{amsfonts}       
\usepackage{nicefrac}       
\usepackage{microtype}      
\usepackage{lipsum}
\usepackage{fancyhdr}       
\usepackage{graphicx}       
\graphicspath{{media/}}     

\title{Tool Wear Segmentation in Blanking Processes with Fully Convolutional Networks based Digital Image Processing}

\author{
  Clemens Schlegel, Dirk Alexander Molitor, Christian Kubik, Daniel Michael Martin, Peter Groche \\
  Institute for Production Engineering and Forming Machines \\
  Technical University of Darmstadt \\
  \texttt{\{clemens.schlegel, dirk.molitor, christian.kubik, daniel.martin, peter.groche\}}\\\texttt{@ptu.tu-darmstadt.de} \\
}

\begin{document}
\maketitle

\begin{abstract}
The extend of tool wear significantly affects blanking processes and has a decisive impact on product quality and productivity. For this reason, numerous scientists have addressed their research to wear monitoring systems in order to identify or even predict critical wear at an early stage. Existing approaches are mainly based on indirect monitoring using time series, which are used to detect critical wear states via thresholds or machine learning models. Nevertheless, differentiation between types of wear phenomena affecting the tool during blanking as well as quantification of worn surfaces is still limited in practice. While time series data provides partial insights into wear occurrence and evolution, direct monitoring techniques utilizing image data offer a more comprehensive perspective and increased robustness when dealing with varying process parameters. However, acquiring and processing this data in real-time is challenging. In particular, high dynamics combined with increasing strokes rates as well as the high dimensionality of image data have so far prevented the development of direct image-based monitoring systems. For this reason, this paper demonstrates how high-resolution images of tools at 600 spm can be captured and subsequently processed using semantic segmentation deep learning algorithms, more precisely Fully Convolutional Networks (FCN). 125,000 images of the tool are taken from successive strokes, and microscope images are captured to investigate the worn surfaces. Based on findings from the microscope images, selected images are labeled pixel by pixel according to their wear condition and used to train a FCN (U-Net).  
\end{abstract}

\keywords{Deep Learning \and Blanking  \and Semantic Segmentation \and Tool Wear \and Transfer Learning}

\section{Introduction}\label{Introduction}
Wear on cutting and blanking tools is one of the main reasons for downtimes and insufficient product quality. Manufacturers are faced with the challenge of determining the optimal intervals to replace or rework their tools, whereby a trade-off problem must be solved between product quality and productivity. Since tool wear depends on a multitude of influencing factors, e.g. lubricant, stroke speed and semi-finished product properties, the accuracy of deterministic prediction of tool wear has shown to be limited \citep{wang2013tool}. For this reason, wear monitoring systems are often based on data-driven models that enable a classification of different wear conditions based on indirect mearusing of the wear state. What these indirect measurement approaches have in common is that additional sensors are integrated into the machine to record different measured variables, which in turn allow conclusions to be drawn about the criticality of the current wear state. \citet{Kubik2022} use force-signals to evaluate cutting edge radii of blanking tools. A 1D CNN is used to classify stamping-induced vibration and acceleration signals in \citep{huang2021stamping}. \citet{Unterberg.2021} process acoustic emission data to classify cutting edge wear in blanking punches and extend their work to evaluate surface roughness on the scrap part resulting from the shearing operation. \citep{Unterberg.2023} Similarly, \citet{ubhayaratne2017audio} use audio signals to evaluate wear and present a signal extracting technique to seperate sound emitted by the stamping operation from surrounding noises. However, this does not directly reveal information about the type and location of wear, which is why indirect monitoring systems do not provide deeper insights into the physics of the process, but need a deeper modeling for signal interpretation. These models are often not available. Additionally, cutting forces and therefore acceleration and acoustic emission are influenced by the properties of the process, such as the tensile strength of the workpiece material \citep{Kubik.2021}. To overcome the limitations of indirect, time series based monitoring systems, direct measurement methods, which record data that contain information about the actual tool wear and reflect the physics of the process more clearly, have to be taken into account.\\
Images are a promising data source as they contain information about the spatial distribution of different types of wear, but face users with the challenge of acquiring and processing images with sufficient resolution. Considering the high stroke rates in blanking processes of up to 1,000 spm \citep{Hirsch.2011} and thus highly dynamic movements of the tool, image-based applications not only place high demands on the integration concept of cameras, exposure and trigger time, but also on real-time and thus stroke-by-stroke processing of the images. First approaches to image-based monitoring of blanking processes circumvent the high process dynamics by using images of the workpieces as a basis for wear and quality monitoring. \citet{molitor2022workpiece} use images of the workpieces and assign them via CNN to 16 punches with different degrees of wear. \citet{kubik2023image} monitor the quality in a high speed blanking process by taking images of the workpieces and extracting features from the fracture surfaces, which can be used to draw conclusions about the workpiece quality.\\
Recently published work on image-based monitoring of wear on drills shows that adaptive contrast enhancement algorithms can extract contours of unworn areas \citep{yu2021machine}. Due to computation times of several seconds, they are unsuitable for stroke-by-stroke monitoring in blanking. This also applies to edge detection algorithms, which are used, for example, for tool wear monitoring on CNC lathes \citep{bagga2021novel}. Other conventional image processing methods, such as different filters, parametric-based models or morphological operation \citep{dutta2013application} require computing times of at least several seconds and are therefore not suitable for inline monitoring. \\
Semantic segmentation approaches based on FCN significantly reduce the required computation times, but have so far only been applied to offline captured images of cutting tools \citep{bergs2020digital} or during intermittent machine downtime by \citet{miao2020u}, which is not feasible for permanent operations in blanking processes. use the U-Net structure as a FCN for real-time processing of cutting tool images and show that less than $200\;\mathrm{ms}$ are required to locate worn areas. Models with such short computing times, in combination with integrated camera systems and finely tuned exposure times, can enable stroke-by-stroke recording and processing of tool images in blanking processes and provide deep insights into the spatial and temporal development of wear. Aside from significantly shorter processing time, \citet{bergs2020digital} showed a superior behavior of machine learning based image segmentation in changing lighting situations compared to tuned traditional computer vision methods.\\
For this reason, this publication presents how a low-cost CMOS camera system is integrated into the interior of a high-speed press and takes images of tools stroke-by-stroke at top dead center. By comparing the images with microscope images, different wear phenomena are located, and random images are subsequently labeled. After applying data augmentation techniques, the labeled images are used to train a hyperparameter-optimised U-Net that classifies pixels into six classes.\\
The paper is organized as follows. 
Section 2 presents the methodology and describes the experimental setup, the camera system used as well as the procedure for labelling the image data and developing the FCN. The training and test results of the modelling as well as a visual comparison between model segmentations and real images are presented in Section 3. Finally, the results are concluded in Section 4.

\begin{figure*}[!b]
    \centering
    \includegraphics[width=\textwidth]{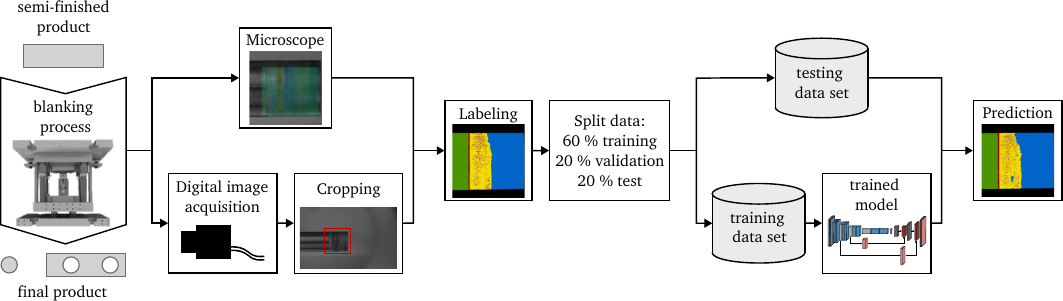}
    \caption{Experimental setup for predicting the wear state of a blanking tool including the data acquisition, the preprocessing, labelling of the dataset and training of the FCN}
    \label{fig:setup}
\end{figure*}

\section{Methodology}\label{sec:Methodology}
\subsection{Experimental setup}\label{sec:Experimental Setup}
The experimental setup to investigate the capabilities of an image-based system for tool wear detection is shown in Fig. \ref{fig:setup}. The camera based data acquisition system is integrated into a blanking process, in which a heat treated cylindrical punch with a diameter of 6 mm is used to cut the final product from a semi-finished product (Table \ref{tab:dc01}). 

The experiments are conducted on a mechanical high-speed press from Bruderer AG (BSTA 810-145) and images are captured during each stroke of the machine. Stroke rates can be selected between 100 spm and $1,000\:\mathrm{spm}$, and the stroke rate used during the tests is set to 600 spm. In each stroke, an image of the blanking tool is taken with a specially developed in situ capable image acquisition system. To label the images for the subsequent training process of the FCN, the tool surface is evaluated by means of white light confocal microscope images ($\mu$surf by NanoFocus). Due to the large amount of time required for microscope imaging, these examinations are performed seven times in the test series. During this test series, a cold-rolled steel DC01 is used (Table \ref{tab:dc01}). It provides a good formability and is expected to induce adhesive wear on the blanking tool. The test series is performed within one day and the machine is stopped only to allow for microscope imaging of the tool and for coil changes.\\

\begin{table}[ht]%
\centering
\caption{Experimental setup}
\begin{tabular}{l c}
\toprule
properties & description\\ \midrule
Material properties& \\
 \;\;\; DC01 & 1.0330\\
 \;\;\; sheet thickness & 1.8 mm \\
 \;\;\; elongation at break & 28 \% \\
 \;\;\; tensile strength & 301.6 MPA \\
 \;\;\; s(tensile strength) & 16.6 MPA \\
 \;\;\; percentage of carbon & 0.12\\
Tool properties &\\ 
 \;\;\; punch diameter & 6 mm\\
 \;\;\; punch hardness & 60-63 HRC\\
 \;\;\; punch material & 1.2379\\
Machine properties &\\
 \;\;\; strokes per minute & 600\\
 \;\;\; stroke length & 35 mm\\\bottomrule     
\end{tabular}
\label{tab:dc01}
\end{table}


\subsection{Data Acquisition}\label{sec:Data Acquisition}
\begin{figure}[h!]
    \centering
    \fontsize{8pt}{9pt}\selectfont
    \def\svgwidth{\columnwidth}
    \begingroup%
  \makeatletter%
  \providecommand\color[2][]{%
    \errmessage{(Inkscape) Color is used for the text in Inkscape, but the package 'color.sty' is not loaded}%
    \renewcommand\color[2][]{}%
  }%
  \providecommand\transparent[1]{%
    \errmessage{(Inkscape) Transparency is used (non-zero) for the text in Inkscape, but the package 'transparent.sty' is not loaded}%
    \renewcommand\transparent[1]{}%
  }%
  \providecommand\rotatebox[2]{#2}%
  \newcommand*\fsize{\dimexpr\f@size pt\relax}%
  \newcommand*\lineheight[1]{\fontsize{\fsize}{#1\fsize}\selectfont}%
  \ifx\svgwidth\undefined%
    \setlength{\unitlength}{777.83436693bp}%
    \ifx\svgscale\undefined%
      \relax%
    \else%
      \setlength{\unitlength}{\unitlength * \real{\svgscale}}%
    \fi%
  \else%
    \setlength{\unitlength}{\svgwidth}%
  \fi%
  \global\let\svgwidth\undefined%
  \global\let\svgscale\undefined%
  \makeatother%
  \begin{picture}(1,0.45350291)%
    \lineheight{1}%
    \setlength\tabcolsep{0pt}%
    \put(0,0){\includegraphics[width=\unitlength,page=1]{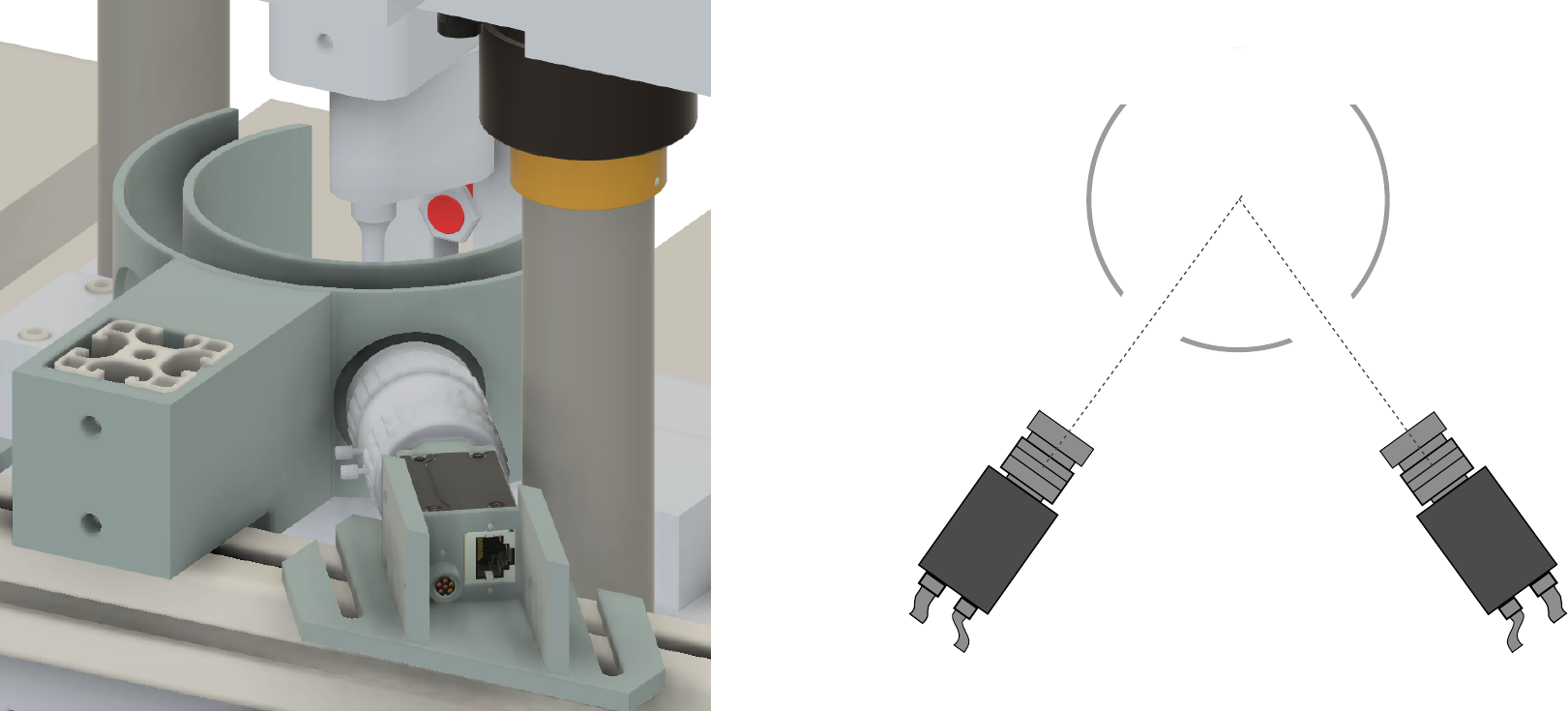}}%
    \put(0.78953765,0.27082187){\color[rgb]{0,0,0}\makebox(0,0)[lt]{\lineheight{1.25}\smash{\begin{tabular}[t]{l}\textit{75}°\end{tabular}}}}%
    \put(0,0){\includegraphics[width=\unitlength,page=2]{Schirm-mit_Foto.pdf}}%
    \put(0.41,0.18755146){\color[rgb]{0,0,0}\makebox(0,0)[lt]{\lineheight{1.25}\smash{\begin{tabular}[t]{l}LED\\mount\\\end{tabular}}}}%
    \put(0,0){\includegraphics[width=\unitlength,page=3]{Schirm-mit_Foto.pdf}}%
    \put(0.4,0.41094277){\color[rgb]{0,0,0}\makebox(0,0)[lt]{\lineheight{1.25}\smash{\begin{tabular}[t]{l}trigger\end{tabular}}}}%
    \put(0,0){\includegraphics[width=\unitlength,page=4]{Schirm-mit_Foto.pdf}}%
    \put(0.45555504,0.03382539){\color[rgb]{0,0,0}\makebox(0,0)[t]{\lineheight{1.25}\smash{\begin{tabular}[t]{c}camera\end{tabular}}}}%
    \put(0,0){\includegraphics[width=\unitlength,page=5]{Schirm-mit_Foto.pdf}}%
    \put(0.39,0.2974434){\color[rgb]{0,0,0}\makebox(0,0)[lt]{\lineheight{1.25}\smash{\begin{tabular}[t]{l}diffusor\end{tabular}}}}%
    \put(0,0){\includegraphics[width=\unitlength,page=6]{Schirm-mit_Foto.pdf}}%
  \end{picture}%
\endgroup%
    \caption{Camera setup for real time image acquisition}
    \label{fig:camera}
\end{figure}

As shown in Fig. \ref{fig:camera}, the image acquisition system consists of industrial cameras (BASLER a2A 1920-51gmPRO) combined with a fixed focal length lens (Ricoh FL-CC2514A-2M). The camera captures monochromatic images with a resolution of $1,920 \times 1,200$ pixels and pixel depth of 8 bit. 
 The main challenge of the in situ camera system is the high stroke rate and thus high speed of the blanking tool during the process. To avoid motion blur, the tool is captured at top dead center of the movement with a short exposure time of the camera. The camera is triggered by an inductive sensor, sensing the upper blanking tool. To allow for short exposure times, a sufficient illumination in combination with a fast aperture of the lens is required. Since the aperture setting additionally determines the depth of focus and therefore is limited in its allowed adjustable range, the light source becomes the main focus of the acquisition system. The lighting system consists of an outer arch carrying the light emitting diodes (LED) and an inner diffuser screen. The diffuser screen from acrylic glass and the circular light arrangement ensure a uniform lighting and cover a total of 210 degrees of the tool. With an exposure time of $50 \; \mu$s, an image acquisition at 600 spm without motion blur is established. After the camera is triggered in the upper dead center, the tool moves $20 \; \mu$m within the duration of the image acquisition of $50 \; \mu$s.


\subsection{Dataset}\label{sec:Pre Processing}
With the image acquisition system described above, a total number of 125,000 images are captured. During the test, wear continuously increased on the peripheral surface of the tool. The images are subsequently used to train the image segmentation model. Training of a neural network requires labeled data, which consists of the input images acquired by the camera system and the desired output, called labels. Training data is chosen as a disjoint subset of the available images. To account for various wear states during the lifecycle of the blanking tool, training images are selected with an equal distance in the number of strokes between labeled images. 309 images of the dataset are chosen for training and testing purposes of the machine learning model. Within this subset, 60 \% of the images are used for training, and 20 \% each for validation and testing of the trained model. 

\begin{figure}[!ht]
\captionsetup[subfigure]{labelformat=empty}
     \centering
     \begin{subfigure}[b]{0.2115\textwidth}
         \centering
         \includegraphics[width=\textwidth]{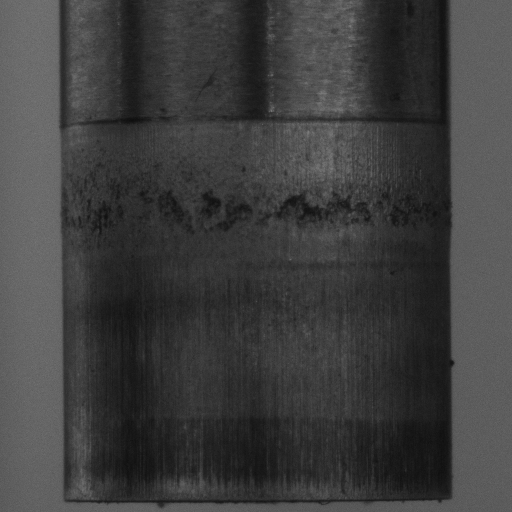}
     \end{subfigure}
     \begin{subfigure}[b]{0.2485\textwidth}
         \centering
         \fontsize{7pt}{9pt}\selectfont
        \def\svgwidth{\columnwidth}
         \begingroup%
  \makeatletter%
  \providecommand\color[2][]{%
    \errmessage{(Inkscape) Color is used for the text in Inkscape, but the package 'color.sty' is not loaded}%
    \renewcommand\color[2][]{}%
  }%
  \providecommand\transparent[1]{%
    \errmessage{(Inkscape) Transparency is used (non-zero) for the text in Inkscape, but the package 'transparent.sty' is not loaded}%
    \renewcommand\transparent[1]{}%
  }%
  \providecommand\rotatebox[2]{#2}%
  \newcommand*\fsize{\dimexpr\f@size pt\relax}%
  \newcommand*\lineheight[1]{\fontsize{\fsize}{#1\fsize}\selectfont}%
  \ifx\svgwidth\undefined%
    \setlength{\unitlength}{198.3893399bp}%
    \ifx\svgscale\undefined%
      \relax%
    \else%
      \setlength{\unitlength}{\unitlength * \real{\svgscale}}%
    \fi%
  \else%
    \setlength{\unitlength}{\svgwidth}%
  \fi%
  \global\let\svgwidth\undefined%
  \global\let\svgscale\undefined%
  \makeatother%
  \begin{picture}(1,0.8525183)%
    \lineheight{1}%
    \setlength\tabcolsep{0pt}%
    \put(0,0){\includegraphics[width=\unitlength,page=1]{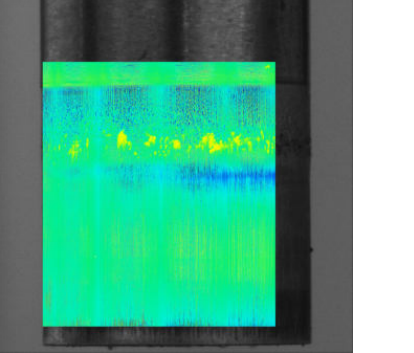}}%
    \put(0.90804217,0.81907305){\color[rgb]{0,0,0}\makebox(0,0)[t]{\lineheight{1.25}\smash{\begin{tabular}[t]{c}$\mu$m\end{tabular}}}}%
    \put(0,0){\includegraphics[width=\unitlength,page=2]{musurf_16_30_new.pdf}}%
    \put(0.96705867,0.25653145){\color[rgb]{0,0,0}\makebox(0,0)[t]{\lineheight{1.25}\smash{\begin{tabular}[t]{c}0\end{tabular}}}}%
    \put(0.97814678,0.3407284){\color[rgb]{0,0,0}\makebox(0,0)[t]{\lineheight{1.25}\smash{\begin{tabular}[t]{c}10\end{tabular}}}}%
    \put(0.97814678,0.42492484){\color[rgb]{0,0,0}\makebox(0,0)[t]{\lineheight{1.25}\smash{\begin{tabular}[t]{c}20\end{tabular}}}}%
    \put(0.97814678,0.50914044){\color[rgb]{0,0,0}\makebox(0,0)[t]{\lineheight{1.25}\smash{\begin{tabular}[t]{c}30\end{tabular}}}}%
    \put(0.97814678,0.59335605){\color[rgb]{0,0,0}\makebox(0,0)[t]{\lineheight{1.25}\smash{\begin{tabular}[t]{c}40\end{tabular}}}}%
    \put(0.97814678,0.67755274){\color[rgb]{0,0,0}\makebox(0,0)[t]{\lineheight{1.25}\smash{\begin{tabular}[t]{c}50\end{tabular}}}}%
    \put(0.97814678,0.76174943){\color[rgb]{0,0,0}\makebox(0,0)[t]{\lineheight{1.25}\smash{\begin{tabular}[t]{c}60\end{tabular}}}}%
    \put(0.96943967,0.17233476){\color[rgb]{0,0,0}\makebox(0,0)[t]{\lineheight{1.25}\smash{\begin{tabular}[t]{c}-10\end{tabular}}}}%
    \put(0.96943967,0.08811915){\color[rgb]{0,0,0}\makebox(0,0)[t]{\lineheight{1.25}\smash{\begin{tabular}[t]{c}-20\end{tabular}}}}%
    \put(0.96943967,0.0039038){\color[rgb]{0,0,0}\makebox(0,0)[t]{\lineheight{1.25}\smash{\begin{tabular}[t]{c}-30\end{tabular}}}}%
  \end{picture}%
\endgroup%
     \end{subfigure}
    \\ \vspace{1mm}
     \centering
     \begin{subfigure}[b]{0.2115\textwidth}
         \centering
         \includegraphics[width=\textwidth, angle=270]{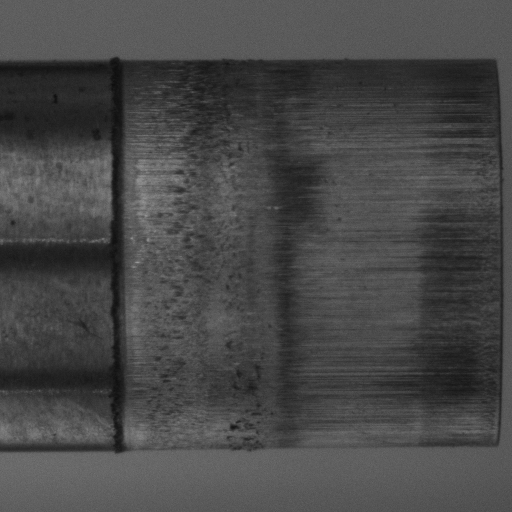}
         \caption{Raw image}
     \end{subfigure}
     \begin{subfigure}[b]{0.2485\textwidth}
         \centering
         \fontsize{7pt}{9pt}\selectfont
        \def\svgwidth{\columnwidth}
         \begingroup%
  \makeatletter%
  \providecommand\color[2][]{%
    \errmessage{(Inkscape) Color is used for the text in Inkscape, but the package 'color.sty' is not loaded}%
    \renewcommand\color[2][]{}%
  }%
  \providecommand\transparent[1]{%
    \errmessage{(Inkscape) Transparency is used (non-zero) for the text in Inkscape, but the package 'transparent.sty' is not loaded}%
    \renewcommand\transparent[1]{}%
  }%
  \providecommand\rotatebox[2]{#2}%
  \newcommand*\fsize{\dimexpr\f@size pt\relax}%
  \newcommand*\lineheight[1]{\fontsize{\fsize}{#1\fsize}\selectfont}%
  \ifx\svgwidth\undefined%
    \setlength{\unitlength}{198.3893399bp}%
    \ifx\svgscale\undefined%
      \relax%
    \else%
      \setlength{\unitlength}{\unitlength * \real{\svgscale}}%
    \fi%
  \else%
    \setlength{\unitlength}{\svgwidth}%
  \fi%
  \global\let\svgwidth\undefined%
  \global\let\svgscale\undefined%
  \makeatother%
  \begin{picture}(1,0.8525183)%
    \lineheight{1}%
    \setlength\tabcolsep{0pt}%
    \put(0,0){\includegraphics[width=\unitlength,page=1]{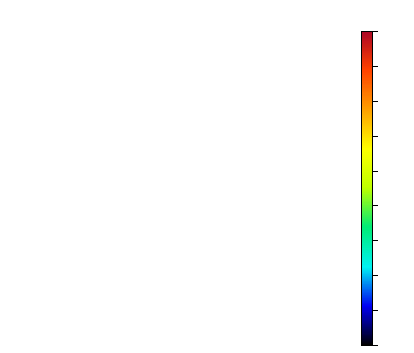}}%
    \put(0.9670597,0.2565357){\color[rgb]{0,0,0}\makebox(0,0)[t]{\lineheight{1.25}\smash{\begin{tabular}[t]{c}0\end{tabular}}}}%
    \put(0.97814787,0.34073256){\color[rgb]{0,0,0}\makebox(0,0)[t]{\lineheight{1.25}\smash{\begin{tabular}[t]{c}10\end{tabular}}}}%
    \put(0.97814787,0.42492921){\color[rgb]{0,0,0}\makebox(0,0)[t]{\lineheight{1.25}\smash{\begin{tabular}[t]{c}20\end{tabular}}}}%
    \put(0.97814787,0.50914472){\color[rgb]{0,0,0}\makebox(0,0)[t]{\lineheight{1.25}\smash{\begin{tabular}[t]{c}30\end{tabular}}}}%
    \put(0.97814787,0.59336022){\color[rgb]{0,0,0}\makebox(0,0)[t]{\lineheight{1.25}\smash{\begin{tabular}[t]{c}40\end{tabular}}}}%
    \put(0.97814787,0.67755698){\color[rgb]{0,0,0}\makebox(0,0)[t]{\lineheight{1.25}\smash{\begin{tabular}[t]{c}50\end{tabular}}}}%
    \put(0.97814787,0.76175373){\color[rgb]{0,0,0}\makebox(0,0)[t]{\lineheight{1.25}\smash{\begin{tabular}[t]{c}60\end{tabular}}}}%
    \put(0.96944072,0.17233905){\color[rgb]{0,0,0}\makebox(0,0)[t]{\lineheight{1.25}\smash{\begin{tabular}[t]{c}-10\end{tabular}}}}%
    \put(0.96944072,0.08812344){\color[rgb]{0,0,0}\makebox(0,0)[t]{\lineheight{1.25}\smash{\begin{tabular}[t]{c}-20\end{tabular}}}}%
    \put(0.96944072,0.00390804){\color[rgb]{0,0,0}\makebox(0,0)[t]{\lineheight{1.25}\smash{\begin{tabular}[t]{c}-30\end{tabular}}}}%
    \put(0,0){\includegraphics[width=\unitlength,page=2]{musurf_18_30_new.pdf}}%
    \put(0.90804221,0.81907305){\color[rgb]{0,0,0}\makebox(0,0)[t]{\lineheight{1.25}\smash{\begin{tabular}[t]{c}$\mu$m\end{tabular}}}}%
  \end{picture}%
\endgroup%

         \caption{microscope image}
     \end{subfigure}
    \caption{Raw images of a blanking tool recorded at 600 spm and
corresponding microscope images}
    \label{fig:microscope}
\end{figure}

\begin{figure}[h!]
\captionsetup[subfigure]{labelformat=empty}
     \centering
     \begin{subfigure}[b]{0.225\textwidth}
         \centering
         \includegraphics[width=\textwidth, angle=270]{18_09_31_,369_2511_rechtscropped.png}
         \caption{Raw image}
     \end{subfigure}
     \begin{subfigure}[b]{0.225\textwidth}
         \centering
         \includegraphics[width=\textwidth, angle=270]{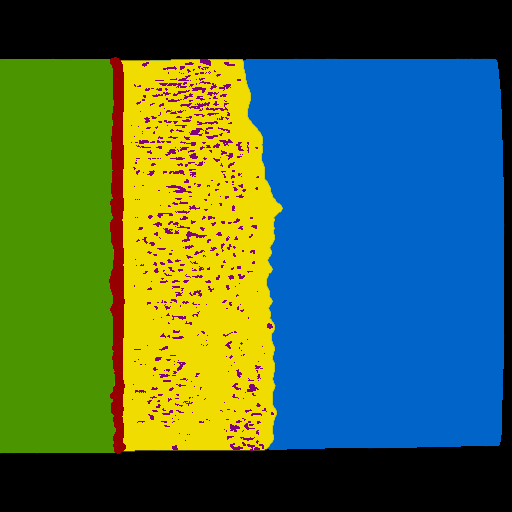}
         \caption{Ground truth}
     \end{subfigure}
    \caption{Raw images of a punch recorded at 600 spm and
corresponding ground truth mask}
    \label{fig:ground truth}
\end{figure}

The label or ground truth mask image is created with the help of the three-dimensional images taken by the white light confocal microscope. In order to assess the different wear phenomena and their appearance on the tool images, the microscope images are projected on the cylindrical surface of the blanking punch. Yellow areas in Fig. \ref{fig:microscope} depict elevated areas, while blue areas show indentations. In the upper part of the tool image, the raw blanking tool can be seen, which does not come in contact with the sheet metal and therefore remains unworn. Underneath the unworn tool area, adhesive wear occurs, resulting in elevations and indentations on the surface. The corresponding area of the blanking tool is in contact with the sheet material in the bottom dead center of the blanking process, resulting in low relative velocity between tool and semi-finished product as well as an alternation of sliding and static friction. Due to high stress on the prominently elevated adhesions during blanking, adhesions detach from the tool and leave behind indentations. The lower part of the punch shows brush marks, which are both visible in the raw images and in the microscope images. Those marks result from abrasive wear and the high relative velocity between the punch and the sheet material while the punch is pushed through and withdrawn from the material. If hardened adhesions detach from the tool, they form grooves in the tool when the tool is withdrawn from the semi-finished product. \citep{Kopp.2017}
After identifying the occurring wear phenomena on the blanking tool, the corresponding ground truth mask for each image of the training and testing subset is created manually. The resulting labels (Table \ref{tab:wear_classes}) provide a pixel-wise segmentation of the image with the following classes and their representation in Fig. \ref{fig:ground truth}.
\begin{table}[ht]%
\centering
\caption{Wear classes and their representation in the ground truth mask in Fig. \ref{fig:ground truth}}
\begin{tabular}{l c}
\toprule
Class name & color in ground truth mask\\ \midrule
background & black \\
unworn area & green \\
contamination & red \\
grooves & blue \\
surface spalling & yellow\\
adhesive wear & violet\\\bottomrule     
\end{tabular}
\label{tab:wear_classes}
\end{table}

An example of a raw image with its corresponding ground truth mask is shown in Fig. \ref{fig:ground truth}. Here, the worn area of the blanking tool is separated in an area of grooves and an area of non-directional surface spalling. According to the microscope images, indentations and elevated areas occur in the area described as surface spalling (yellow) in the ground truth mask. As described, the elevations correspond to adhesions and the indentations are likely to result from detached adhesions. In the corresponding raw image of the blanking tool, both wear patterns display as small dark areas and can not be further separated. Therefore, all small dark areas or dots are labeled as adhesive wear (purple). During the process, contamination builds up due to abrasive wear between tool and semi-finished product. This contamination is shown as red area in the ground truth mask. The vertical grooves described earlier are shown in blue color.

\subsection{Data Augmentation}\label{sec:Data Augmentation}
\begin{figure}[h!]
\captionsetup[subfigure]{labelformat=empty}
     \centering
     \begin{subfigure}[b]{0.23\textwidth}
         \centering
         \includegraphics[width=\textwidth, angle=270]{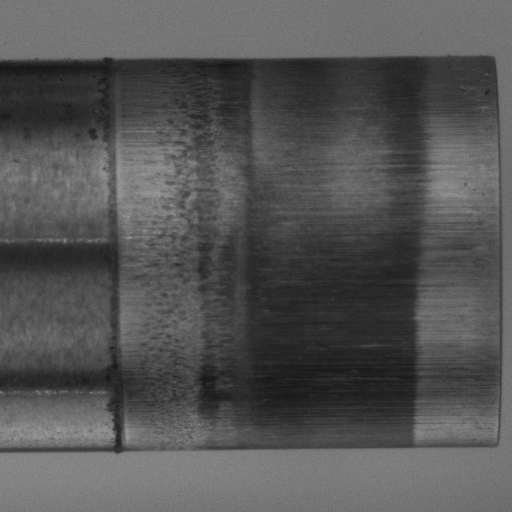}
         \caption{Original}
     \end{subfigure}
     \begin{subfigure}[b]{0.23\textwidth}
         \centering
         \includegraphics[width=\textwidth, angle=270]{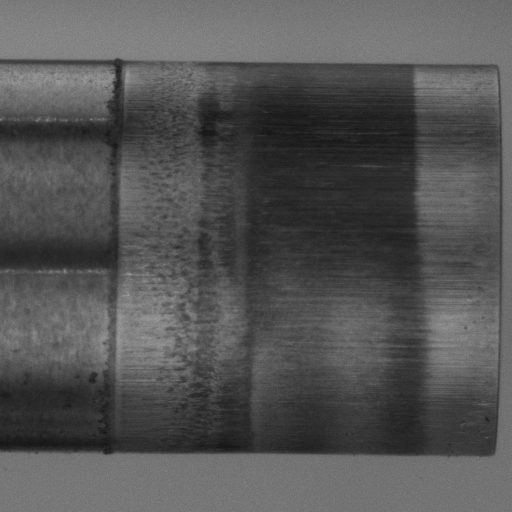}
         \caption{Horizontal flip}
     \end{subfigure}
     \\
     \begin{subfigure}[b]{0.23\textwidth}
         \centering
         \includegraphics[width=\textwidth, angle=270]{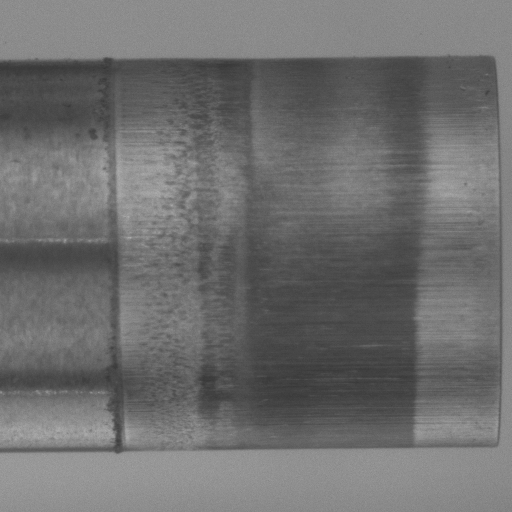}
         \caption{Random brightness}
     \end{subfigure}
     \begin{subfigure}[b]{0.23\textwidth}
         \centering
         \includegraphics[width=\textwidth, angle=270]{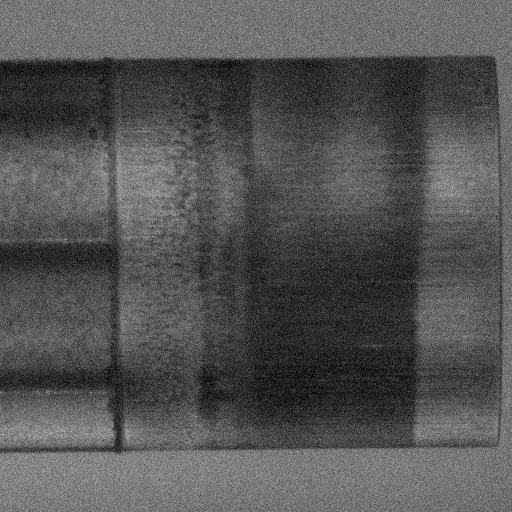}
         \caption{Gaussian Noise}
     \end{subfigure}
        \caption{Examplary data augmentation techniques used in this work}
        \label{fig:data_aug}
\end{figure}

The training and accordingly the performance of image-based DL algorithms depend on the quality and amount of the available annotated data. \citep{DirkAlexanderMolitor.2021} Since the generation of labels is often costly and time-consuming, data augmentation (DA) techniques can be used to generate artificial datasets based on the captured tool images. DA for image processing could be achieved through geometrical transformations or a shift in exposure and color space, or by using generative adversarial networks. \citep{Shorten.2019} However, those networks do not necessarily improve training further compared to regular DA techniques while being computationally more expensive. \citep{Mertes.2020}
Accordingly, the regular augmentation techniques listed in Table \ref{tab:data_aug} are applied to the training dataset.
\begin{table}[ht]%
\centering
\caption{Data Augmentation applied to the training dataset}
\begin{tabular}{l c}
\toprule
Manipulation technique & Settings\\ \midrule
Horizontal flip & probability $p=0,5$\\
Gamma value  & $\gamma\in[0,8; 1,2]$\\
Contrast multiplier  & $1 \pm 0,2$\\
Brightness multiplier  & $1 \pm 0,2$\\
Gaussian noise  & $X^\mathrm{GN}_{i,j,k}\in[-50; 50]$\\ \bottomrule     
\end{tabular}
\label{tab:data_aug}
\end{table}
Each image of the dataset is augmented two times, which leads to a total training dataset of 555 images. Exemplary image manipulations are shown in Fig. \ref{fig:data_aug}. While the differences introduced by the DA techniques are barely perceptible to the naked eye, they significantly contribute to improving the generalizability of the trained model.

\subsection{Modelling}\label{sec:Modelling}
\begin{figure*}[ht]
	\begin{center}
		\centering
		\includegraphics[width=\textwidth]{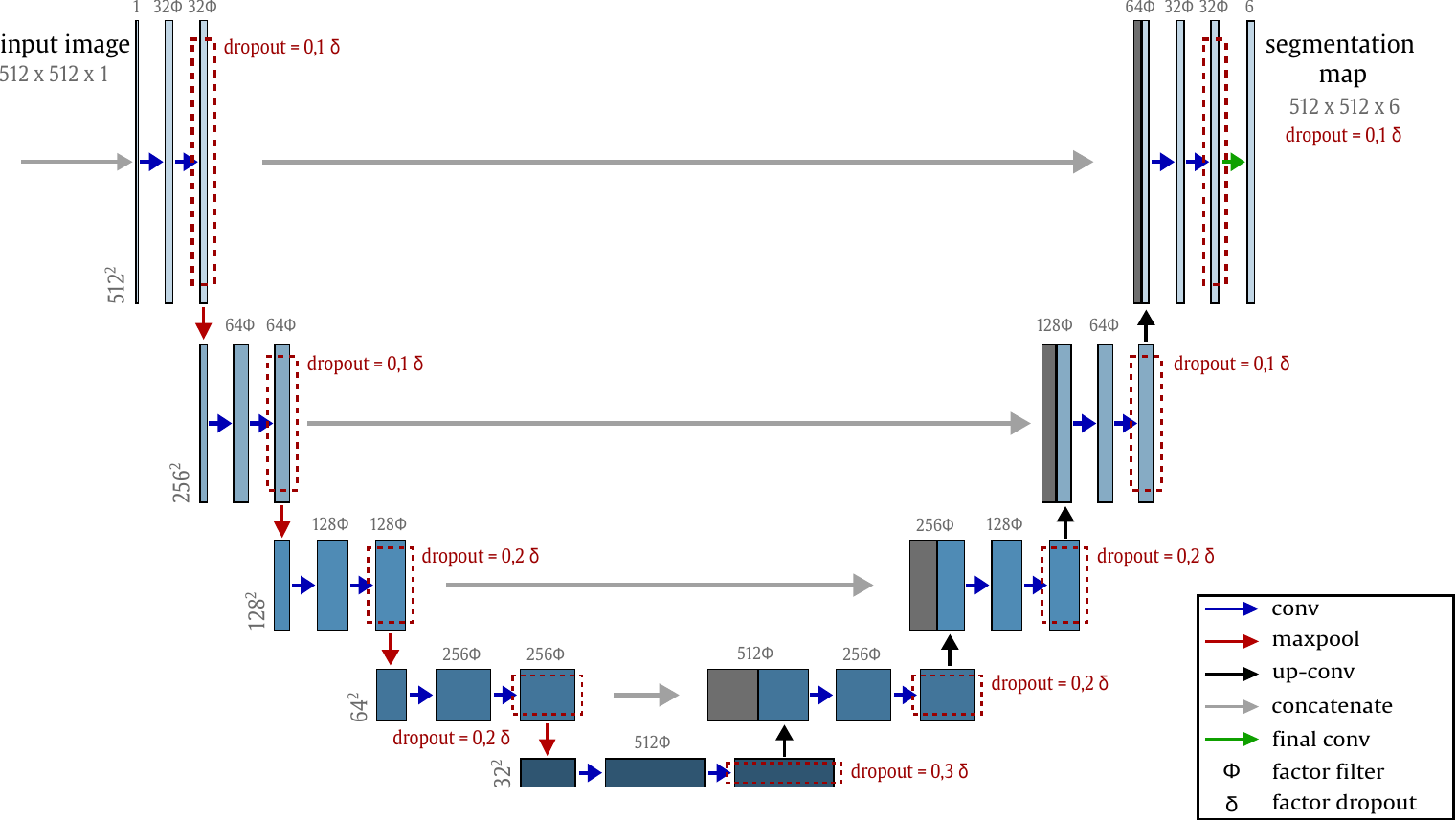}
		\caption{Adapted U-Net architecture with adaptable dropout and filter layers}
		\label{fig:unet}
	\end{center}
\end{figure*}

The semantic segmentation of the wear on the blanking tool is achieved using a FCN. The model architecture of FCN is based on convolutional neural networks (CNN), which enables image-to-image processing by outputting a segmentation heatmap based on an input-image \citep{Long.2014}. The resulting architecture consists of a spatial contraction (encoder) and expansion component (decoder), where the encoder is equivalent to a CNN. In the present work, a U-Net, as presented by \citet{Ronneberger.2015}, is used. In previous work, U-Net-based approaches have been used in comparable applications such as wear segmentation on cutting tools. \citep{bergs2020digital}. \citet{Huang.2019} evaluated different FCN approaches and their performance within a texture recognition task. Here, U-Net showed high prediction accuracy, while being one of the fastest algorithms. Short computation time for segmentation is particularly relevant for in situ wear detection, with segmentations of up to 10 images per second. As shown in Fig. \ref{fig:unet}, the U-Net architecture consists of multiple different operations and is characterized by an encoder and a decoder path.

In Fig. \ref{fig:unet}, multichannel feature maps of the input image are shown as blue boxes with a given size and number of feature channels. In each step of the encoder path, two consecutive $3 \times 3$ convolutions are applied, followed by a Rectified Linear Unit (ReLU) activation. Then, a $2 \times 2$ max pooling operation is performed. The pooling operation acts on each channel of the feature map individually and reduces the xy-size of the feature map by propagating the maximum value of each $2\times2$ window to the feature map in the next step. The number of filters in the convolutional layer doubles with each step of the U-Net, which means that spacial expansion is halved with each encoder step, while the number of feature channels doubles. Feature channels further down the encoder path contain a higher level of information about the content of the image, while no spatial context is retained. In the following decoder path, the spatial information gets restored to form the segmentation map, which indicates the content of the image and where it could be seen in the image. The decoder path consists of an upsampling and two subsequent convolutional layers. Each of the convolutional layers includes a ReLu activation function. Finally, similar to the original FCN model \citep{Long.2014}, a $1\times1$ convolution per class is used to transform the result of the last $3\times3$ convolutional layer into the desired heatmap form. This last convolutional layer has a softmax activation function. 

To adapt the size of the network to the dataset in a later optimization step, a variable number of filters is provided in each convolutional layer of the U-Net. For this purpose, a factor $\Phi$ is used, which is multiplied with the respective filter number of the U-Net from \citet{Ronneberger.2015} and results in the model sizes shown in Table \ref{tab:model_size}. 
\begin{table}[ht]%
\centering
\caption{Model size in dependence of the scaling parameter $\Phi$}
\begin{tabular}{l c c}
\toprule
$\Phi$ & number of parameters & memory size\\ \midrule
0.0625 & 121,678 & 1.6 MB\\
0.125 & 485,718 & 5.6 MB\\
0.25 & 1,940,902 & 22.4 MB\\
0.5 & 7,759,686 & 89 MB\\
1 & 31,030,918 & 335.3 MB\\\bottomrule     
\end{tabular}
\label{tab:model_size}
\end{table}

Additionally, an adaptable dropout rate is used to help generalization of the trained network. Here, the factor $\delta$ is multiplied with the dropout rate used by \citet{Harrison.2021} and optimized in the following training of the U-Net.

\subsection{Evaluation Metric for Semantic Segmentation}\label{sec:Evaluation}
\begin{figure}[h!]
    \centering
    \fontsize{9pt}{9pt}\selectfont
    \def\svgwidth{0.6\columnwidth}
    \begingroup%
  \makeatletter%
  \providecommand\color[2][]{%
    \errmessage{(Inkscape) Color is used for the text in Inkscape, but the package 'color.sty' is not loaded}%
    \renewcommand\color[2][]{}%
  }%
  \providecommand\transparent[1]{%
    \errmessage{(Inkscape) Transparency is used (non-zero) for the text in Inkscape, but the package 'transparent.sty' is not loaded}%
    \renewcommand\transparent[1]{}%
  }%
  \providecommand\rotatebox[2]{#2}%
  \newcommand*\fsize{\dimexpr\f@size pt\relax}%
  \newcommand*\lineheight[1]{\fontsize{\fsize}{#1\fsize}\selectfont}%
  \ifx\svgwidth\undefined%
    \setlength{\unitlength}{379.68814757bp}%
    \ifx\svgscale\undefined%
      \relax%
    \else%
      \setlength{\unitlength}{\unitlength * \real{\svgscale}}%
    \fi%
  \else%
    \setlength{\unitlength}{\svgwidth}%
  \fi%
  \global\let\svgwidth\undefined%
  \global\let\svgscale\undefined%
  \makeatother%
  \begin{picture}(1,0.27070608)%
    \lineheight{1}%
    \setlength\tabcolsep{0pt}%
    \put(0,0){\includegraphics[width=\unitlength,page=1]{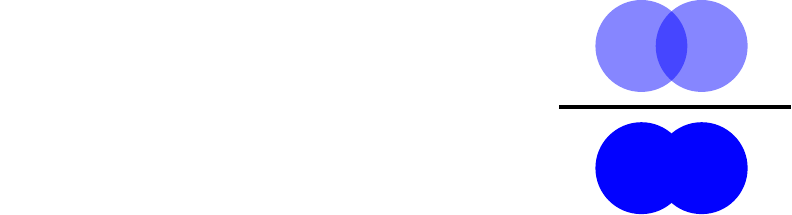}}%
    \put(0.6646267,0.10979954){\color[rgb]{0,0,0}\makebox(0,0)[t]{\lineheight{1.25}\smash{\begin{tabular}[t]{c}=\end{tabular}}}}%
    \put(0.16683961,0.10979954){\color[rgb]{0,0,0}\makebox(0,0)[t]{\lineheight{1.25}\smash{\begin{tabular}[t]{c}=\end{tabular}}}}%
    \put(0,0){\includegraphics[width=\unitlength,page=2]{IoU.pdf}}%
    \put(0.42622954,0.15910282){\color[rgb]{0,0,0}\makebox(0,0)[t]{\lineheight{1.25}\smash{\begin{tabular}[t]{c}Intersect \end{tabular}}}}%
    \put(0.42549141,0.05910334){\color[rgb]{0,0,0}\makebox(0,0)[t]{\lineheight{1.25}\smash{\begin{tabular}[t]{c}Union \end{tabular}}}}%
    \put(0.07291656,0.1091031){\color[rgb]{0,0,0}\makebox(0,0)[t]{\lineheight{1.25}\smash{\begin{tabular}[t]{c}IoU \end{tabular}}}}%
  \end{picture}%
\endgroup%
    \caption{Equation and graphical representation of the evaluation metric IoU}
    \label{fig:iou}
\end{figure}

An appropriate evaluation metric is crucial to assess the quality of the neural network's prediction. Therefore, the pixelwise classification identified by the FCN is compared to the ground truth mask of the test data. In this work, the accuracy metric Intersect over Union (IoU), also known as Jaccard index, is used to compare ground truth and prediction. \citep{Jaccard.1912} The metric compares the commonly labeled pixels of ground truth and predicted mask with the union of both masks and therefore presents a relative accuracy value (Fig. \ref{fig:iou}).

\section{Training and Test Results}\label{sec:Results}
The following section describes training and hyperparameter optimization, as well as results of the trained networks.

\subsection{Training Results and Hyperparameter Optimization}\label{sec:Training Results}
During the training of neural networks, numerous parameters are adjusted. 
The success of the training heavily depends on the chosen hyperparameters, making the optimization of these hyperparameters an essential part of developing a neural network.
The network's ability to segment individual classes cannot be solely evaluated by the accuracy of the validation dataset's predictions. Instead, the IoUs of the individual classes can be used to determine how well each class has been learned. 
The objective of the optimization is to achieve a stable learning process that generates high values for the mean IoU and the IoU of the class \textit{adhesive wear}. 

The first optimization approach tests different filter numbers in the convolutional layers and various batch sizes, as these significantly influence the size of the model and the backpropagation frequency and therefore the speed of training. The goal of this first optimization approach is to find a configuration with a high batch size and a low kernel number to enable further optimization quickly and efficiently. 
To investigate different model sizes, the scaling parameter $\Phi$ is multiplied with the filter numbers of the original U-Net as shown in Table \ref{tab:model_size}. In addition, different batch sizes are examined to further reduce training time. After 500 episodes, the combination of batch size 4 and a kernel count of 25 \% of the original U-Net shows the best results (Table \ref{tab:opti1}).
\begin{table}[ht]%
\centering
\caption{Maximum mean IoU of different parameter configurations during the hyperparameter optimization}
\begin{tabular}{l l c c c c c}
\toprule
 \multicolumn{2}{c }{}& \multicolumn{5}{c}{batch size}\\
 \multicolumn{2}{c }{}& 1&2&4&8&16\\ \midrule
\parbox[t]{1mm}{\multirow{4}{*}{\rotatebox[origin=c]{90}{Factor $\Phi$}}} & 1/16 & 0,767 & 0,768 & 0,765 & 0,758 & 0,611\\
&1/8 & 0,777 & 0,783 & 0,774 & 0,776 & 0,777\\
&1/4 & 0,858 & 0,873 & \textbf{0,877} & 0,844 & 0,827\\
&1/2 & 0,855 & 0,833 & 0,865 & 0,871  & 0,873\\\bottomrule     
\end{tabular}
\label{tab:opti1}
\end{table}

\begin{figure}
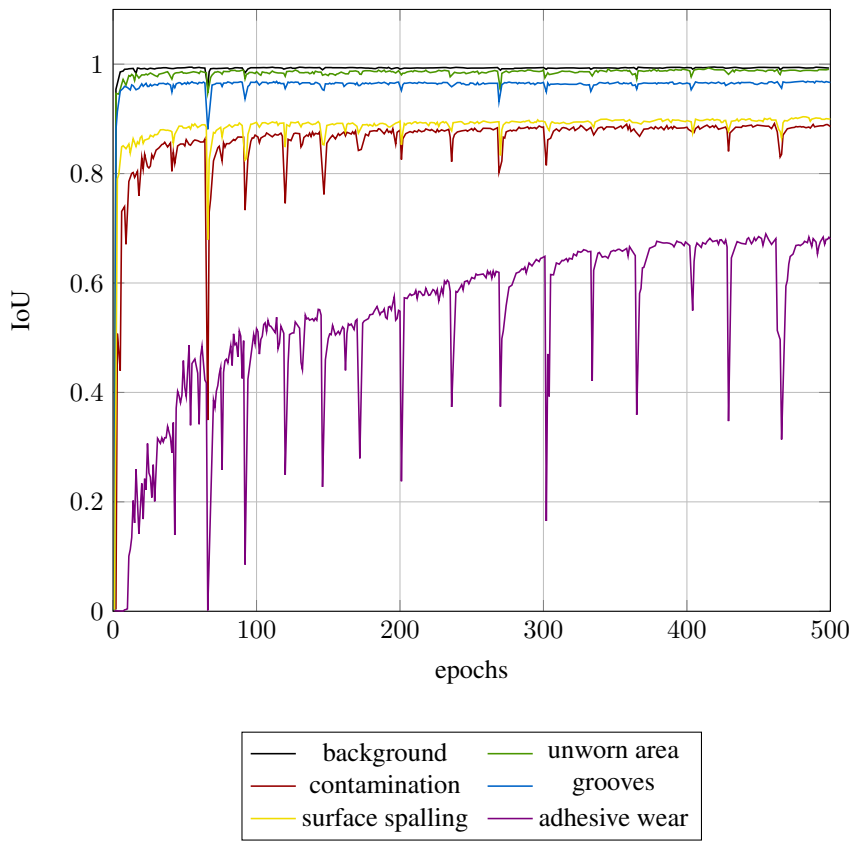

\centering

\caption{Training curves over 500 epochs with $\Phi=0.25$ and batch size = 4}
\label{fig:opt1_plot}
\end{figure}

Figure \ref{fig:opt1_plot} shows the IoUs of the individual classes learned by the model over the course of 500 epochs. While Fig. \ref{fig:opt1_plot} shows successful improvement in all classes, significant drops are observed in the IoU of class \textit{adhesive wear} during the course of training. These drops are also observed, to a lesser extent, in the trends of the other IoUs. To achieve a more stable training, different learning rates and dropout rates are examined. Lower learning rates lead to a slower but more stable learning progress and the amount of drops in the IoU during training can be reduced significantly. However, learning rate must be chosen carefully, since a particularly low learning rate prevents the learning of classes, that are represented on fewer pixels. During optimization, a learning rate of $5.4 \cdot 10^{-4}$ is chosen. Different dropout rates influence the generalizability of the trained model. A low dropout rate leads to overfitting and the validation accuracy falls behind the training accuracy. Similarly to the learning rate, a high dropout rate results in poor learning of the lesser represented classes such as the adhesive wear. During optimization, a factor for the dropout rate of $\delta=0.48$ is chosen. With these parameters, the IoU of the class adhesive wear could be improved to 0.7123 after 500 epochs with a mean IoU of 0.9064.
Figure \ref{fig:opt1_plot} shows that the classes are learned to varying degrees of success and especially the start of learning varies between the classes. Individual classes are represented to different extents in the dataset, and less represented classes are learned more slowly and less effectively. Accordingly, an attempt will be made to promote the learning of the important but less represented class of adhesive wear by using an additional weighting of this class. Sample weights are assigned to each pixel of the input image and Bayesian optimization is used to determine optimal weights for the classes adhesive wear and surface spalling. The remaining classes do not receive additional weighting. Higher class weights lead to an improved performance in the first epochs. However, using weights in the optimization process can lead to unstable training, with higher weights generally reducing training stability. This results in a lower IoU of 0.6716 for class adhesive wear. In a further optimization approach, the use of class weights is combined with a lower learning rate, as a lower learning rate has proven to stabilize training. This approach is successful compared to the previous optimization of the class weights. Training is stabilized and the critical IoU of class adhesive wear could be improved to 0.6962. However, the results still do not exceed the training without class weights. In a last attempt, the performance of the model could be enhanced by continuing the training of the previously best model. Building on the results of previous optimizations, a slight emphasis on the class adhesive wear by using class weights is introduced. However, care is taken to avoid introducing instabilities through the additional weighting in the training process. The IoU of the class adhesive wear could be improved to 0.7645 within 500 epochs of training. Consequently, the model resulting from the last optimization attempt was used for the final evaluation of the developed monitoring system.

\subsection{Test Results}\label{sec:Test Results}

\begin{figure}[h!]
\captionsetup[subfigure]{labelformat=empty}
     \centering
     \begin{subfigure}[b]{0.225\textwidth}
         \centering
         \includegraphics[width=\textwidth, angle=270]{18_09_31_,369_2511_rechtscropped_label.png}
         \caption{Ground truth}
     \end{subfigure}
     \begin{subfigure}[b]{0.225\textwidth}
         \centering
         \includegraphics[width=\textwidth, angle=270]{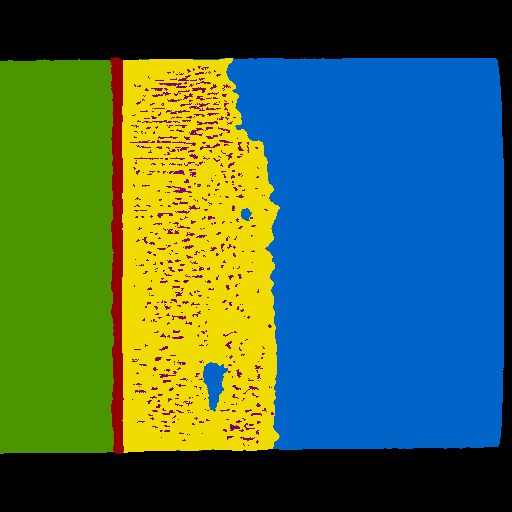}
         \caption{Prediction}
     \end{subfigure}
    \caption{Comparison of real images and deep learning segmentation}
    \label{fig:segmentation comparison}
\end{figure}

Figure \ref{fig:segmentation comparison} shows a comparison between the image segmentation produced by the trained FCN model and the ground truth determined from the microscope image displayed in Fig. \ref{fig:microscope}. The image depicted has not been included in the training set of the FCN, and as such, remains unobserved by the model prior to testing. The prediction shows high agreement with the respective ground truth with slightly lower accuracy for the classes \textit{surface spalling}, \textit{contamination} and \textit{adhesive wear}. 

As shown in Table \ref{tab:results_IoU}, the IoU values for these classes are slightly lower compared two the remaining classes \textit{background}, \textit{unworn area} and \textit{grooves}. 

\begin{figure}[t]
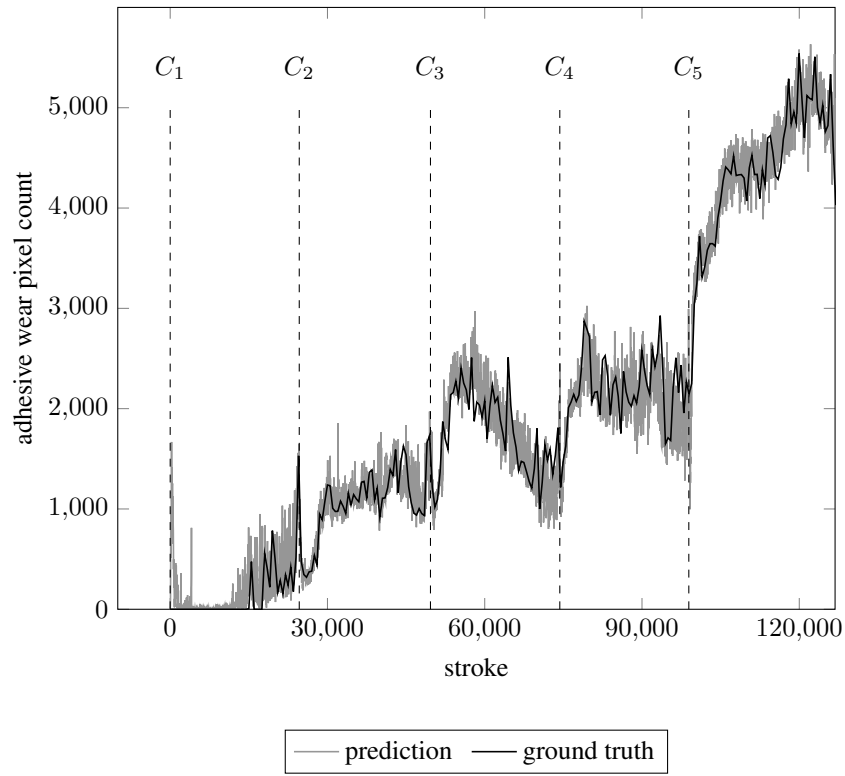

\centering

\caption{Pixel count for the class adhesive wear over the full data set: Comparison between ground truth and prediction}
\label{fig:pixel_count}
\end{figure}

\begin{table*}[t]%
\centering
\caption{Accuracy metrics for each predicted class and mean value in training, validation and test data set}
\begin{tabular*}{\textwidth}{l @{\extracolsep{\fill}} ccc}
\toprule
 & training & validation & test \\ \midrule
 IoU of class \textit{background} (black) & 0.9924 & 0.9927 & 0.9887 \\
 IoU of class \textit{unworn area} (green) & 0.9955 & 0.9876 & 0.9886 \\
 IoU of class \textit{contamination} (red) & 0.9729 & 0.8926 & 0.8800 \\
 IoU of class \textit{grooves} (blue) & 0.9921 & 0.8926 & 0.9714 \\
 IoU of class \textit{surface spalling} (yellow) & 0.9675 & 0.9016 & 0.9185 \\
 IoU of class \textit{adhesive wear} (violet) & 0.6873 & 0.7645 & 0.7696 \\ \midrule
 Mean IoU & 0.9346 & 0.9172 & 0.9195 \\ \bottomrule
\end{tabular*}
\label{tab:results_IoU}
\end{table*}
The deviation between prediction and ground truth for the class surface spalling results from the low sharpness of separation for class grooves in the camera images. There is no objective criterion that allows for a clear separation between the mentioned areas, and a subjective selection of the areas is made. The differences in IoUs between individual classes result from the characteristics of each area. Areas with a large ratio of area perimeter to area tend to achieve lower IoUs, since slight displacements of the prediction compared to the ground truth result in a high relative number of misclassified pixels. Mean IoU and IoUs of the training, validation, and test sets differ, particularly for adhesions, with the test set showing higher IoU than the train or validation sets. The lower IoU for training data may be due to excessive deviation resulting from DA, or due to the random split and limited number of training samples leading to an underrepresentation of certain features. 
To further diversify the evaluation of the trained network, another comparison method is introduced. The method is based on comparing the pixel count of individual classes in the image. Unlike in the IoU, small shifts in the prediction compared to the original label in the image are not considered errors. As the exact position of the adhesive wear is not necessarily relevant for the evaluation of wear, a temporal relationship of wear development can be simplified. To allow for a thorough evaluation of the trained model, a prediction is made for $10\;\%$ of all acquired images and compared to the ground truth of training and test data set. The comparison of prediction and ground truth in Fig. \ref{fig:pixel_count} shows, that the predicted pixel count for the class adhesive wear follows the ground truth over the full data set. Only at the beginning of the experiment, the predicted pixel count deviates from the ground truth. At this early stage, adhesive wear has not yet occurred on the blanking tool. Therefore very little training data is available and the FCN does not learn to correctly segment unworn tools. 

Additionally, the pixel count in Fig. \ref{fig:pixel_count} shows repeated dips over the course of the experiment. Those can be explained by the varying process conditions due to microscope imaging and change of coils ($C_i$) of the semi-finished product. Both processes are carried out simultaneously. While cleaning the blanking tool carefully before the microscope imaging is necessary to evaluate wear phenomena, the acquisition condition, i.e. the appearance of adhesive wear, changed significantly due to this step. After the stops denoted as $C_i$ in Fig. \ref{fig:pixel_count}, the pixel count for the class rises significantly again.



\section{Conclusion}\label{sec:Conclusion}
Most monitoring approaches for blanking processes focus on indirect measurement approaches to assess quality and tool condition. However, direct monitoring the blanking tool can provide a more nuanced understanding of the various types of wear phenomena that occur. The approach presented here is capable of capturing images of the blanking tool during the process at a rate of 600 spm and classifying tool wear to quantify the worn areas. To enable image acquisition during high-speed blanking processes, lighting conditions have been optimized to provide even and bright illumination of the tool. By capturing images at top dead center with an exposure time of $t=50\;\mathrm{\mu s}$ seconds, sharp images can be obtained without motion blur, even at rates of up to 10 images per second. The wear detection approach uses a U-Net to segment the different wear classes, achieving a promising mean IoU of 0.9195 and an IoU of 0.7696 for the most relevant class \textit{adhesive wear} on the test data. Training was performed with a total number of 555 augmented images generated from 185 raw images. The training data covers a wide range of wear conditions that occur during the life of a blanking tool. Data preparation for the segmentation network is crucial and ground truth masks have to be made with great care. While microscope images help to create the ground truth maps correctly, the resulting interruptions of the process alter the captured images significantly. The appearance of adhesive wear changes, and the segmented prediction, while agreeing with the corresponding ground truth, may not show the actual signs of wear on the tool. Hence, a duality arises regarding whether to increase the number of microscope images to accurately identify the wear pattern or decrease the number of images to maintain consistent imaging conditions.

The promising segmentation results can be applied for tool condition monitoring or for predicting remaining useful tool life. Based on the presented work, new possibilities for future research arise. In a first step, similar models should be evaluated on different semi-finished products and punch geometries.
To ensure the in situ capabilities of the model in even faster and more complicated applications, different network architectures should be considered. The optimization of network parameters in this work showed, that smaller models could be trained faster and more efficiently. Therefore considering a small network to only segment the most relevant wear phenomena would allow the prediction to be done on less powerful engines in-line. Additionally more recent U-Net-based architectures like U-Net++  \citep{Zhou.2019} or U-Net 3+ \citep{Huang2020_UNet3+} could be used. These architectures improve prediction accuracy by adding nested and dense skip connections between the encoder and decoder path. To segment the occurring wear phenomena more accurately, depth information could be captured by means of stereo vision technique. Resulting images could be evaluated with a 3D-U-Net \citep{Cicek.2016}. Labeled data of high speed processes could be further used to create and train models based on indirect measurements and time series signals. Thereby, a holistic representation of the machine and tool condition could be established.

\section*{Acknowledgements}
The results of this paper are achieved within the
project “ProKI – Demonstrations- und Transfernetzwerk KI für die Umformtechnik ProKI-Darmstadt” funded by the German Federal Ministry of Education and Research (BMBF). The authors wish to thank for funding and supporting this project. Furthermore, the authors sincerely thank Bruderer AG for providing the high-speed press BSTA 810-145.
\section*{Conflict of interest}
The authors declare that they have no conflict of interest.

\end{document}